\documentclass[a4paper]{article}

\usepackage{INTERSPEECH2019}
\usepackage{subcaption}
\usepackage{xcolor}
\usepackage{amsmath}
\usepackage{hyperref}

\title{Affective Conditioning on Hierarchical Attention Networks applied to Depression Detection from Transcribed Clinical Interviews}
\name{Danai Xezonaki$^1$, Georgios Paraskevopoulos$^{1,3}$, Alexandros Potamianos$^{1,2,3}$, Shrikanth Narayanan$^{2,3}$}

\address{
$^1$School of ECE, National Technical University of Athens, Athens, Greece\\
$^2$Signal Analysis and Interpretation Laboratory (SAIL), USC, Los Angeles, CA, USA\\
$^3$Behavioral Signal Technologies, Los Angeles, CA, USA
}
\email{dxezonaki@gmail.com, geopar@central.ntua.gr, potam@central.ntua.gr, shri@sipi.usc.edu}

\begin{document}

\maketitle
\begin{abstract}

\end{abstract}
In this work we propose a machine learning model for depression detection from transcribed clinical interviews. 
Depression is a mental disorder that impacts not only the subject’s mood but also the use of language.
To this end we use a Hierarchical Attention Network to classify interviews of depressed subjects. 
We augment the attention layer of our model with a conditioning mechanism on linguistic features, extracted from affective lexica.
Our analysis shows that individuals diagnosed with depression use affective language to a greater extent than not-depressed.
Our experiments show that external affective information improves the performance of the proposed architecture in the General Psychotherapy Corpus and the DAIC-WoZ 2017 depression datasets, achieving state-of-the-art $71.6$ and $68,6$ F1 scores respectively.


\section{Introduction}
Depression is a serious mood disorder that affects the way people think and behave.
According to WHO~\cite{who}, it is estimated that over $300$ million people suffer from depression, which corresponds to the $4.4\%$ of the world’s population. 
Indicative symptoms of depression can be the loss of interest in everyday activities, sleeping and eating disorders, feelings of worthlessness, sadness and exhaustion, or even thoughts of suicide~\cite{dsm}. 
WHO also states that over $800{,}000$ suicide deaths are reported each year due to depression, while for $15$-$29$-year-old people, it is the leading factor of death.
The growing amount of available online data opens opportunities to perform data driven analyses and develop computational algorithms to assist specialists in the field of psychology, study depression and refine clinical methods and protocols.

Depression detection is the problem of identifying signs of depression in individuals. 
These signs might be identified in peoples' speech, facial expressions and in the use of language.
In our task, we consider the binary classification task of detecting depression in transcribed clinical sessions between a therapist and a client. 
These sessions provide valuable insights of the cognitive and behavioral functioning of clients. 
Therefore, we leverage behavioral and psycholinguistic cues of the client and therapist language to enhance our models.

Previous studies have shown that depression affects the language use of depressed individuals. 
They tend to use more absolutist words~\cite{absolute}, negatively valenced-words and the pronoun ``I''~\cite{essays} and mention pharmaceutical treatment for depressive disorder~\cite{social,drugs}.
People in distress also make less use of first person plural pronouns~\cite{ZIMMERMANN} and become more self-focused~\cite{selfattention}.
In~\cite{metadata}, linguistic metadata features are employed across with external knowledge including domain-adapted lexica while in~\cite{deprlexica}, Losada et al. propose evaluation methods of existing depression lexica and create sub-lexica based on part-of-speech tagging.
Moreover, for the General Psychotherapy Corpus~\footnote{\href{http://alexanderstreet.com}{http://alexanderstreet.com}}, Malandrakis et al.~\cite{samedataset} have explored differences in language between therapist and client using psycholinguistic norms and Imel et al.~\cite{samedataset1} have identified semantic topics discussed in therapy sessions.
Other studies based on therapy sessions have also predicted empathy through motivational interviews~\cite{james} and have explored behavioral coding learning models for different psychotherapy approaches~\cite{Gibson}.

Hierarchical models have been proposed for document classification tasks, in order to leverage the hierarchies existing in the document structure and construct a document-level representation based on turn-level and word-level representations~\cite{hierarchical2}.
These models have been augmented with attention mechanisms~\cite{attention1,attention2} to identify salient words and sentences in the document~\cite{hierarchical}.
In addition, affective lexica have been published~\cite{liwc, bingliu, afinn, emolex, semeval, mpqa} which can effectively contribute in sentiment analysis. 
As a useful external linguistic knowledge, they can be incorporated into neural architectures~\cite{deprcondition}.
In~\cite{conditioning}, attentional conditioning methods were proven to enhance model performance for sentiment classification tasks.

In this work we focus on the problem of depression detection in psychotherapy sessions.
We employ a two-staged hierarchical network functioning at word and turn-level. 
Each level is equipped with an attention mechanism to extract important content from different parts of the session. 
To leverage the affective context of depressive language we employ a conditioning method~\cite{conditioning} using affective lexica and fuse them in the word-level attention network.
We also incorporate the summary attributed to each session into the proposed architectures.
Our key contribution is that 
we integrate existing affective information which improves the results of our hierarchical neural network for depression detection, especially in the case we have small amount of data.
This fact results in high performing models and improved robustness across two corpora.

\section{Methodology}
\label{methodology}
Our task is a document classification task, where the input to the model is the transcription of the therapy session and the output is a prediction of the subjects depression status. Hierarchical Neural Networks are a natural fit for document classification, since sessions are composed of turns, which consist of words, forming a hierarchical textual structure.

\subsection{Hierarchical Model}
\begin{figure}[ht]
  \centering
  \includegraphics[width=\linewidth]{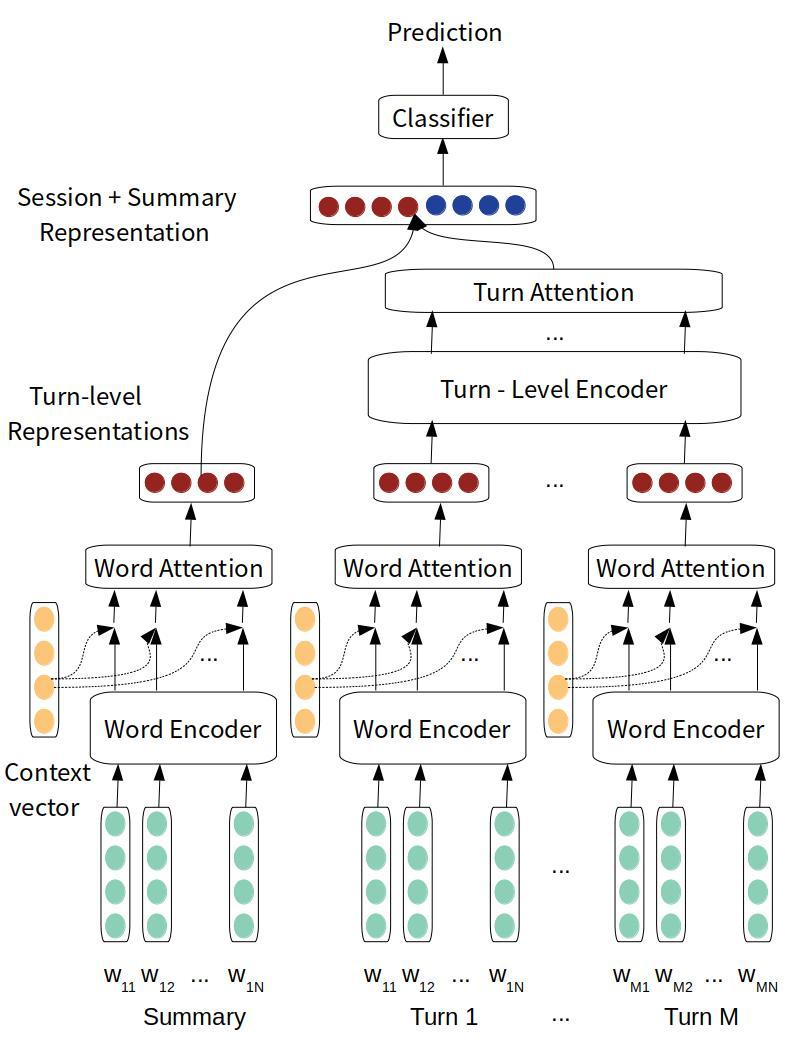}
  \caption{Hierarchical Model with Attentional Conditioning}
  \label{fig:hier_basic}
\end{figure}

The input sequence of words are embedded into a low-dimensional vector space. In document classification, we want to extract the hierarchies existing in documents in a bottom-up manner. To this end, we use a two-stage hierarchical network that operates at word and turn-level, as we can see in Fig.~\ref{fig:hier_basic}. Both the word-level and the turn-level encoders are implemented using Recurrent Neural Networks (RNN). Since not all words or turns contribute equally to the final session representation, we augment both encoders with an attetion mechanism~\cite{attention1}.
At the first level of the hierarchy, a word-level encoder produces turn-level reprepresentations. We feed the words of each turn to the encoder and then combine them to a single representation using an attention mechanism. Let $h_{ki}$ be the annotation of the $i$-th word in the $k$-th turn obtained through the word-level encoder. The k-th turn representation results as follows:
%
%
%
\begin{align}
\begin{split}
  \gamma_{ki} &= g(h_{ki}), \\
  \alpha_{ki} &= \frac{e^{\gamma_{ki}}}{\sum_i{e^{\gamma_{ki}}}}, \\
  t_k &= \sum_i \alpha_{ki} \cdot h_{ki}
\end{split}
  \label{eq1}
\end{align}

where $g$ is a learnable mapping, $a_{ki}$ are the attention weights for each word and $t_k$ is the $k$-th turn representation.

The session representations are extracted in a similar manner. The turn representations $t_k$ are fed into the turn-level  encoder and then the attention weights are calculated. The final representations are the weighted sum of the turn-level encoder hidden states with the attention weights.
\begin{align}
\begin{split}
  \beta_k &= f(t_k), \\
  \tau_k &= \frac{e^{\beta_{k}}}{\sum_i{e^{\beta_{k}}}}, \\
  r &= \sum_k \tau_k \cdot \beta_k
\end{split}
  \label{eq2}
\end{align}
%

%
where $f$ is a learnable mapping, $\tau_k$ are the attention weights and $r$ is the session-level representation.

\subsection{External knowledge conditioning}

According to~\cite{essays,deprlexica}, the affective content can be a distinguishing factor between depressed and not-depressed language. 
Based on this observation, we employ external linguistic knowledge about the affective content of words.
These features can be obtained by sources created by human experts.
We consider emotion, sentiment, valence and psycho-linguistic annotations for words. 
Specifically, we construct a context vector $c_{ki}$ for each word $i$ in turn $k$, where each dimension corresponds to an annotation from existing affective lexica. We set missing dimensions to zero and we integrate the context vector in the attention mechanism of the word-level encoder.
Specifically, we concatentate, $||$, the context vector to the hidden representation of each word $h_{ki}$, modifying Eq.~\ref{eq1}:

\begin{align}
\begin{split}
  \gamma_{ki} &= g(h_{ki}||c_{ki}), \\
  \alpha_{ki} &= \frac{e^{\gamma_{ki}}}{\sum_i{e^{\gamma_{ki}}}}, \\
  t_k &= \sum_i \alpha_{ki} \cdot (h_{ki}||c_{ki})
\end{split}
  \label{eq3}
\end{align}
Eq.~\ref{eq3} shows that we compute the intermediate representations $\gamma_{ki}$ using both the word hidden states and the context vector. 
The softmax function is then applied to $\gamma_{ki}$ to create the attention weights distribution $\alpha_{ki}$.
The incorporation of external information at this level can force higher values for attention weights corresponding to salient affective words.
We also use the concatenated $h_{ki}||c_{ki}$ to create the turn representations $t_k$ to propagate the affective features to the turn-level encoder.

\subsection{Integrating Session Summary}
In the General Psychotherapy Corpus, sessions are accompanied with a summary given by an expert. 
This summary can be seen as a high-level overview of the topics discussed during the session and is denoted as ``title'' in the dataset. 
Similar to~\cite{title}, we extract the summary’s vector representation through the word-level encoder and concatenate it directly with the final session representation, before feeding it to the classifier.
Let $o_t$ be the summary representation obtained through the word-level encoder. Concatenating it with the session representation (~\ref{eq2}) produces the final vector ($o_t || r$) that is fed to the classifier.

\begin{figure*}[ht]
  \centering
  \begin{subfigure}{.48\textwidth}
  \includegraphics[width=\linewidth]{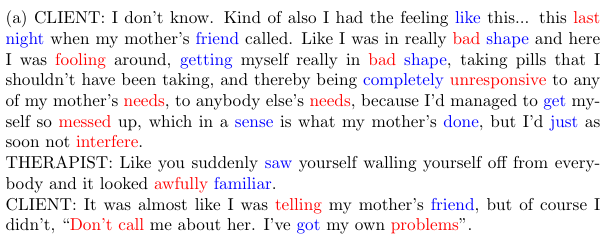}
  \label{fig:example-affected}
  \end{subfigure}
  \begin{subfigure}{.48\textwidth}
  \includegraphics[width=\linewidth]{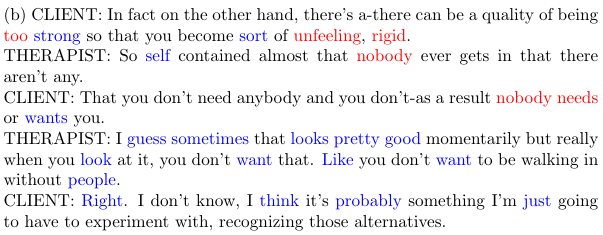}
  \label{fig:example-non-affected}
  \end{subfigure}
  \vspace{-1\baselineskip}
  \caption{Example of sessions for (a) a depressed client and (b) a not-depressed client. Blue: positive words, Red: negative words, as found in LIWC, AFINN and Bing Liu Opinion Lexicon.}
    \label{fig:example}
\end{figure*}

\section{Data Overview and Analysis}
\noindent \textbf{General Psychotherapy Corpus:} We use the \textit{General Psychotherapy Corpus} (GPC) by the ``Alexander Street Press''. This dataset contains over $1{,}300$ transcribed therapy sessions, which cover a variety of clinical approaches. Metadata are also provided at session level and include demographic information for both therapist and client, the symptoms that the clients are experiencing and a summary attributed to each session, labeled as ``title''.
As some of the sessions are conducted with more than two speakers, we extract a subset of $1{,}262$ sessions which consist of one therapist and one client. Each session is comprised of consecutive dialogue turns, annotated as therapist-side or client-side turns. Among the total of $1{,}262$ sessions, $881$ of them are annotated as ``not-depressed'' samples whereas the rest $381$ are annotated as ``depressed''.

\vspace{0.2cm}

\noindent \textbf{DAIC-WoZ:} The DAIC-WoZ dataset is part of the DAIC corpus~\cite{distress}. 
It contains a set of clinical interviews which were carried out so as to assist the task of detecting distress disorders.
The interview is conducted between a client and a virtual agent serving as the therapist, called Ellie, which is controlled by a human interviewer placed in another location~\cite{simsensei}. 
The dataset contains audio and video recordings as well as the transcripts of clinical interviews.
Data are split into train, development and test set, consisting of $107$, $35$, $47$ samples respectively and depression is evaluated on the PHQ-8 depression scale. 

\subsection{Data Analysis on the General Psychotherapy Corpus}
In this section, we explore statistics regarding the language used by depressed and not-depressed individuals in the GPC corpus.
As mentioned in Section~\ref{methodology}, sessions are provided as consecutive turns, as shown in Fig.~\ref{fig:example}.
We see that the depressed client's language contains more negative affective content.
In Table~\ref{tab:table1}, we present the average number of tokens in turns of clients and therapists. We notice that the clients speak twice as much as the therapists on average.

\begin{table}[th]
  \caption{Dialogue turns statistics for therapists and clients}
  \label{tab:table1}
  \centering
  \begin{tabular}{ cc }
    \toprule
    \textbf{Features} & \textbf{Sum} \\
    \midrule
    Average number of turns/session & $196$\\
    Average number of tokens in turns & $32.3$\\
    Average number of tokens in client turns & $42.9$\\
    Average number of tokens in therapist turns & $20.7$\\
    \bottomrule
  \end{tabular}
\end{table}

Next, we compare the use of language between depressed and not-depressed clients. 
In particular, we are interested in the use of words that express positive and negative sentiment, sadness and anxiety.
To this end, we employ the LIWC lexicon~\cite{liwc}, which provides psycho-linguistic annotations for $18{,}504$ words, for $73$ different word categories.
In Table~\ref{tab:table2}, we compare the vocabularies of depressed and not-depressed people and specifically show the vocabulary sizes and the percentage of their words which are associated with these four affective word categories, in LIWC.
We see that depressed subjects use a more consice vocabulary, but include a higher percentage of affective words in it.
This hints to the importance of incorporating knowledge about affective language in depression detection.

\begin{table}[th]
  \caption{Vocabulary use statistics between the two classes}
  \label{tab:table2}
  \centering
  \begin{tabular}{ccc}
    \toprule
    \textbf{Features} & \textbf{Depressed} & \textbf{Not-depressed} \\
    \midrule
    Samples & $381$ & $881$\\
    Total turns & $41589$ & $88191$\\
    Vocabulary size & $16166$ & $23201$\\
    Number of affective words & $1672$ & $2036$ \\
    Percentage of affective words & $10.34\%$ & $8.77\%$\\
    \bottomrule
  \end{tabular}
\end{table}

Moreover, in Table~\ref{tab:table3}, we present the percentages of the four word categories in the language of clients.
Specifically, we split the dataset into the samples of depressed and not-depressed people. 
Subsequently, we use the LIWC lexicon and count the number of occurences of words that belong to each of these affective categories.
Finally, we compute their percentages, in the total words of the samples of depressed and not-depressed people.
The results indicate that depressed individuals tend to use more negatively-valenced words, which stands in agreement with the related literature~\cite{essays}.   

\begin{table}[th]
  \caption{Percentage of occurences of affective word categories in client language across the two classes}
  \label{tab:table3}
  \centering
  \begin{tabular}{ccc}
    \toprule
    \textbf{Categories} & \textbf{Depressed} & \textbf{Not-depressed}\\
    \midrule
    Positive sentiment&  $2.17$\% & $2.26$\%\\
    Negative sentiment&  $1.38$\% & $1.30$\%\\
    Sadness &  $0.32$\% & $0.32$\%\\
    Anxiety &  $0.30$\% & $0.22$\%\\
    \bottomrule
  \end{tabular}
\end{table}

\section{Experimental Setup}
\label{experimental_setup}
We employ five network architectures. 
As a weak baseline model we employ Tf-Idf for feature extraction and an SVM classifier with linear kernel (SVM). 
Moreover, we develop a Hierarchical Attention-based Network with no external knowledge, which is referred to as HAN. 
Subsequently, we augment this model with affective conditioning at the attention mechanism, where lexicon annotations are concatenated with word hidden states before the word-level attention layer (HAN+L). 
We also utilize the session summaries that are provided with the GPC dataset and extend the HAN model with the integration of the summary's representation before the classification layer (HAN+S). 
Our last model results from the combination of the two previous network architectures (HAN+L+S). 
As there is no summary assigned to the sessions of the DAIC-WoZ corpus, we evaluate the HAN and HAN+L models on this dataset. 
For our experiments on GPC we report macro-averaged F1 score due to the class imbalance present in the datasets. This score is calculated using $5$-fold cross-validation, where each fold contains an $80\%-20\%$ train-validation split of the original data.
In the case of DAIC-WoZ we also measure Unweighted Average Recall (UAR) as in~\cite{same} and present results on the development set.

\vspace{0.1cm}

\noindent \textbf{Lexical features:} Lexical representations for words are extracted from six affective lexica, namely LIWC~\cite{liwc}, Bing Liu Opinion Lexicon~\cite{bingliu}, AFINN~\cite{afinn}, Subjectivity Lexicon (MPQA)~\cite{mpqa}, SemEval 2015 English Twitter Lexicon (Semeval15)~\cite{semeval} and NRC Emotion Lexicon (Emolex)~\cite{emolex}.
AFINN, Semeval15 and Bing Liu provide $1D$ positive/negative sentiment annotations for $6{,}786$, $1{,}515$ and $2{,}477$ words respectively. 
MPQA provides $4D$ sentiment ratings for $6{,}886$ words.
Emolex provides $19D$ emotion ratings for $14{,}182$ words.
LIWC provides $73D$ psycholinguistic annotations for $18{,}504$ words.
The combined six lexica provide a vocabulary coverage of $25{,}534$ words.
These features are concatenated into a $99$-dimensional context vector. 

\vspace{0.1cm}

\noindent \textbf{Data preprocessing:} 
To preprocess the data of the GPC, we keep only the dialogue turns of the therapist and the client by removing speaker tags and any extra information provided as notes in the transcript. 
Next, for both corpora we tokenize the speaker turns by splitting them into words.
We use $300D$ GloVe~\cite{glove} pretrained word embeddings, trained on the Common Crawl corpus, to extract word representations.

\vspace{0.1cm}

\noindent \textbf{Implementation details:}  Our models consist of two encoder layers, where a Bi-directional Gated Recurrent Unit (GRU) is implemented on each stage. 
All encoders use $300$ hidden size and $0.2$ dropout rate. 
Model parameters are optimized using Adam with $10^{-3}$ learning rate.
Models are trained for a maximum of $40$ epochs and we use early stopping to select the model with the lowest validation loss. 
Models are implemented using Pytorch framework~\cite{pytorch}.

\section{Results and Discussion}
We compare the performance of the proposed models when given as input the client turns (Client), the therapist turns (Therapist) or the whole dialogue (Client+Therapist).
In Table~\ref{tab:table4} we present the results for the GPC dataset.
We see that the integration of external affective and psycholinguistic features improves model performance for all model configurations over the HAN and SVM baselines.
Furthermore, we notice that when we add the summary information we also gain a performance boost, sometimes greater that the external affective information.
Summary and lexica integration leads to a performance increase when we provide only the client data.
In addition, we see that the client turns are more important for depression detection, as expected, and incorporation of the therapist turns contributes little to the overall model performance.
Based on our results, the best performance can be achieved by the HAN+L+S model, while HAN+L model performs best if such annotation is not available.

\begin{table}[th]
  \caption{Results of different architectures on the GPC}
  \label{tab:table4}
  \centering
  \begin{tabular}{cccc}
    \toprule
    \multicolumn{1}{c}{\textbf{Experiment}} & 
    \multicolumn{1}{c}{\textbf{Client}} &
    \multicolumn{1}{c}{\textbf{Therapist}} &
    \multicolumn{1}{c}{\textbf{Client+Therapist}} \\
    \midrule
    SVM & $0.478$ & $0.464$ & $0.484$\\
    HAN & $0.681$ & $0.647$ & $0.695$\\
    HAN+S & $0.698$ & $0.641$ & $\mathbf{0.718}$\\
    HAN+L & $0.693$ & $\mathbf{0.659}$ & $0.706$\\
    HAN+L+S & $\mathbf{0.715}$ & $0.640$ & $\mathbf{0.716}$\\
    \bottomrule
  \end{tabular}
\end{table}

In Table~\ref{tab:table5} we present results for the DAIC-WoZ dataset. We observe that affective conditioning significantly improves the performance over the baseline model (HAN). Our HAN+L model also leads to improved F1 and AUR scores over the models proposed in~\cite{same}. Overall, we see that conditioning of external psycholinguistic knowledge in this small dataset (189 samples) enhances the performance and the results are comparable to these of the GPC dataset.

\begin{table}[th]
  \caption{Results of the DAIC-WoZ corpus development set}
  \label{tab:table5}
  \centering
  \begin{tabular}{ccc}
    \toprule
    \multicolumn{1}{c}{\textbf{Method}} & 
    \multicolumn{1}{c}{\textbf{F1-macro}} & 
    \multicolumn{1}{c}{\textbf{UAR}} \\
    \midrule
    \cite{same} HCAN& $0.51$ & $0.54$\\
    \midrule
    \cite{same} HLGAN& $0.60$ & $0.60$\\
    \midrule
    HAN & $0.47$ & $0.54$\\
    \midrule
    HAN+L& $\mathbf{0.69}$ & $\mathbf{0.72}$\\
    
    \bottomrule
  \end{tabular}
\end{table}

\section{Conclusions and Future Work}
We propose a novel model for depression detection with integrated external affective and psycholinguistic information.
Our model is a Hierarchical Attention Network that encodes words and dialogue turns in different levels of the architecture.
The external features are integrated into the attention mechanism, forcing the attention weights to focus on salient affective information.
The external knowledge integration leads to high performing models and improved robustness for both the small datasets ($1{,}262$ and $189$ samples respectively) we explore.
In the future, we plan to extend our architecture to model the dialogue interaction of the therapist and the client.
Finally, we plan to incorporate more elaborate information sources, e.g. expert knowledge bases from psychologists.

\section{Acknowledgements}
We would like to thank psychologists Evangelia Prassopoulou and Anastasios Panopoulos who gave us a thorough insight into Depressive disorder and their psychological approach on the treatment.

\bibliographystyle{IEEEtran}

\bibliography{mybib}


\end{document}